\pdfoutput=1

\documentclass[11pt]{article}

\usepackage[final]{acl}

\usepackage{times}
\usepackage{latexsym}

\usepackage[T1]{fontenc}

\usepackage[utf8]{inputenc}

\usepackage{microtype}

\usepackage{inconsolata}
\usepackage{xspace}
\usepackage{makecell}
\usepackage{cleveref}
\usepackage{wrapfig}
\usepackage{paralist}
\usepackage{graphicx}
\usepackage{booktabs}
\usepackage{multirow}
\usepackage{makecell}
\usepackage{subcaption}
\usepackage{enumitem}
\usepackage{float}
\usepackage{stfloats}

\usepackage{listings}
\usepackage{xcolor}
\definecolor{green}{HTML}{33A02C}
\definecolor{red}{HTML}{E31A1C}

\usepackage{pifont}
\newcommand{\cmark}{\ding{51}}%
\newcommand{\xmark}{\ding{55}}%

\renewcommand{\cite}{\citep}

\newcommand{\eg}{\emph{e.g.,}\xspace}

\newcommand{\topmost}{\textsc{TopMost}\xspace}

\crefname{section}{Sec.}{Sec.}

\title{Towards the \textsc{TopMost}: A Topic Modeling System Toolkit}

\author{
    Xiaobao Wu \qquad Fengjun Pan \qquad Anh Tuan Luu \\
    Nanyang Technological University \\
    \texttt{xiaobao002@e.ntu.edu.sg} \quad \texttt{fengjun001@e.ntu.edu.sg} \quad \texttt{anhtuan.luu@ntu.edu.sg}
}

\begin{document}
\maketitle
\begin{abstract}
    Topic models have a rich history with various applications and have recently been reinvigorated by neural topic modeling.
    However, these numerous topic models adopt totally distinct datasets, implementations, and evaluations.
    This impedes quick utilization and fair comparisons, and thereby hinders their research progress and applications.
    To tackle this challenge, we in this paper propose a Topic Modeling System Toolkit (\topmost).
    Compared to existing toolkits, \topmost stands out by supporting more extensive features.
    It covers a broader spectrum of topic modeling scenarios with their complete lifecycles, including datasets, preprocessing, models, training, and evaluations.
    Thanks to its highly cohesive and decoupled modular design, \topmost enables rapid utilization, fair comparisons, and flexible extensions of diverse cutting-edge topic models.
    These improvements position \topmost as a valuable resource to accelerate the research and applications of topic models.
        Our code, tutorials, and documentation are available at {\url{https://github.com/bobxwu/topmost}}.
        Our demo video is at \url{https://youtu.be/9bN-rs4Gu3E?si=LunquCRhBZwyd1Xg}.
\end{abstract}

\section{Introduction}
    Topic models have been a fundamental and prevalent research area for decades. They aim to understand documents in an unsupervised fashion by discovering latent topics from them and inferring their topic distributions \cite{churchill2022evolution}.
    Besides the basic topic modeling scenario \cite{blei2003latent},
    various other scenarios have been explored, \eg hierarchical, dynamic, and cross-lingual topic modeling \cite{griffiths2003hierarchical,blei2006dynamic,mimno2009polylingual}.
    Current topic models can be categorized into two types.
    The first type is conventional topic models which follow
    either 
    non-negative matrix factorization \cite{lee2000algorithms,kim2015simultaneous,shi2018short}
    or
    probabilistic graphical models via Markov Chain Monte Carlo \cite{steyvers2007probabilistic} or Variational Inference \cite{blei2017variational}.
    The second type is recently popular neural topic models, learned through gradient backpropagation~\cite{zhao2021topic,wu2023survey}.
    Thus they can avoid the laborious model-specific derivations of conventional models, attracting more research attention.
    Due to the effectiveness and interpretability of topic models,
    they have inspired various downstream tasks and applications,
    \eg text analysis and content recommendation \cite{boyd2017applications}.
    Despite these significant achievements, quick utilization and fair comparisons of various topic models remain a formidable challenge.
    The reason lies in their unsystematic model implementations as well as inconsistent dataset and evaluation settings across papers, even within a paper \cite{hoyle2021is}.

\begin{table}
    \centering
         \setlength{\tabcolsep}{0.5mm}
        \renewcommand{\arraystretch}{1.15}
        \resizebox{\linewidth}{!}{
        \begin{tabular}{llcc}
        \toprule
        \multicolumn{2}{c}{Topic Modeling Scenario} & OCTIS & \textbf{\topmost} \\
        \midrule
        \multirow{3}[0]{*}{Basic topic modeling} & Datasets & \textcolor{green}{\cmark} & \textcolor{green}{\cmark} \\
              & Models & \textcolor{green}{\cmark} & \textcolor{green}{\cmark} \\
              & Evaluations & \textcolor{green}{\cmark} & \textcolor{green}{\cmark} \\
        \midrule
        \multirow{3}[0]{*}{Hierarchical topic modeling} & Datasets & \textcolor{green}{\cmark} & \textcolor{green}{\cmark} \\
              & Models & \textcolor{green}{\cmark} & \textcolor{green}{\cmark} \\
              & Evaluations & \textcolor{red}{\xmark} & \textcolor{green}{\cmark} \\
        \midrule
        \multirow{3}[0]{*}{Dynamic topic modeling} & Datasets & \textcolor{red}{\xmark} & \textcolor{green}{\cmark} \\
              & Models & \textcolor{red}{\xmark} & \textcolor{green}{\cmark} \\
              & Evaluations & \textcolor{red}{\xmark} & \textcolor{green}{\cmark} \\
        \midrule
        \multirow{3}[0]{*}{Cross-lingual topic modeling} & Datasets & \textcolor{red}{\xmark} & \textcolor{green}{\cmark} \\
              & Models & \textcolor{red}{\xmark} & \textcolor{green}{\cmark} \\
              & Evaluations & \textcolor{red}{\xmark} & \textcolor{green}{\cmark} \\
        \bottomrule
        \end{tabular}%
        }
        \caption{
            Comparison between the latest OCTIS \cite{terragni2021octis} and \topmost.
            Our \topmost covers more topic modeling scenarios and their corresponding datasets, models, and evaluations.
        }
        \label{tab_comparison}%
\end{table}%

\begin{figure}[!ht]
    \centering
    \includegraphics[width=\linewidth]{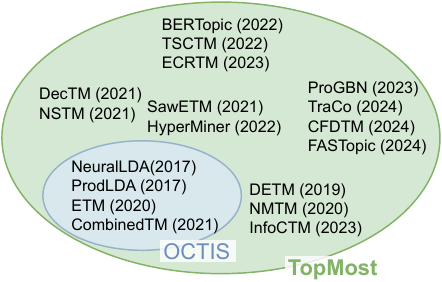}
    \caption{
        Comparison of neural topic models in OCTIS and our \topmost.
        Our \topmost covers more latest neural topic models than OCTIS.
    }
    \label{fig_venn_comparison}
\end{figure}

    Several topic modeling toolkits emerge in response to this challenge by integrating different topic models and evaluations.
    However, they fail to fully meet practical requirements due to lacking certain essential features.
    Early toolkits \cite{mccallum2002mallet,rehurek2011gensim,qiang2020short,lisena2020tomodapi}
    often lack the support for neural topic models or necessary steps in the topic modeling lifecycle, \eg data preprocessing and evaluations.
    The latest toolkit OCTIS \cite{terragni2021octis} is more comprehensive, but as shown in \Cref{tab_comparison} and \Cref{fig_venn_comparison}, it solely considers basic and hierarchical topic modeling scenarios and overlooks the latest advancements of neural topic models, offering only two neural topic models introduced after 2018.
    As a consequence, these issues pose hurdles to the comparisons, developments, and applications of topic models.

    To resolve these issues,
    we in this paper introduce \textbf{Top}ic \textbf{Mo}deling \textbf{S}ystem \textbf{T}oolkit (\textbf{\topmost}),
    which supports extensive features.
    In contrast to existing toolkits, \topmost thoroughly incorporates the most prevalent topic modeling scenarios: \textbf{basic, hierarchical, dynamic, and cross-lingual topic modeling}, as well as the latest neural topic models as detailed in \Cref{tab_comparison} and \Cref{fig_venn_comparison}.
    It covers the entire lifecycles of these scenarios, including datasets, preprocessing, models, training, and evaluations.
    More importantly, \topmost adheres to an object-oriented paradigm with a highly cohesive and decoupled modular design.
    This enhances the readability and extensibility of \topmost,
    enabling users to flexibly customize their own datasets, models, and evaluations for their diverse research or application purposes.
    As a result, \topmost excels in fulfilling the practical requirements of topic modeling.
    We conclude the advantages of our \topmost as follows:
    \begin{itemize}[leftmargin=*,parsep=0pt]
        \item
            \topmost provides handy and complete cutting-edge topic models for various scenarios;
        \item
            \topmost allows users to effortlessly and fairly compare topic models through comprehensive evaluation metrics;
        \item
            \topmost with better readability and extensibility facilitates the smooth development of new topic models and downstream applications.
    \end{itemize}

\begin{figure}
    \centering
    \includegraphics[width=0.95\linewidth]{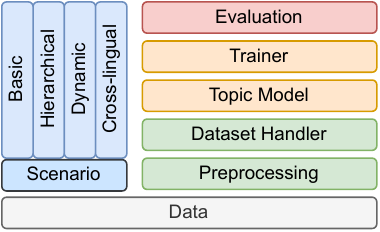}
    \caption{
        Overall architecture of \topmost.
        It covers the most common topic modeling scenarios and decouples data loading, model constructions, model training and evaluations in topic modeling lifecycles.
    }
    \label{fig_architecture}
\end{figure}

\section{Related Work}
    Throughout the long history of topic modeling, numerous toolkits have emerged and gained widespread adoption.
    The earliest among those include
    Mallet \cite{mccallum2002mallet}~\footnote{\url{https://mimno.github.io/Mallet/topics.html}}
    and gensim \cite{rehurek2011gensim}~\footnote{\url{https://radimrehurek.com/gensim/}}.
    While these fundamental frameworks sufficiently embrace conventional topic models, they generally overlook the recent advancements in neural topic models.
    STTM \cite{qiang2018sttm} particularly focuses on probabilistic short text topic models, like BTM \cite{yan2013biterm} and DMM \cite{yin2014dirichlet}.
    A more recent entrant, OCTIS \cite{terragni2021octis}, integrates both conventional and neural topic models.
    Nevertheless, it merely covers basic and hierarchical topic modeling scenarios and neglects the latest neural topic models developed after 2018.
    Moreover, OCTIS couples the implementations of model construction and training, exacerbating the challenges of toolkit maintenance.
    Different from these existing work, our \topmost extensively incorporates a spectrum of popular topic modeling scenarios and the latest developments in neural topic models.
    In addition, our \topmost clearly decouples each step (data, models, and training) in the topic modeling lifecycles, resulting in neat code structures and simplified maintenance.

\begin{table*}
    \centering
    \small
     \renewcommand{\arraystretch}{1.2}
    \begin{tabular}{l|l|l|l}
    \toprule
    \textbf{Topic Modeling Scenarios} & \textbf{Topic Models} & {\textbf{Evaluation Metrics}} & {\textbf{Datasets}} \\
    \midrule
    Basic topic modeling
    & {\makecell[l]{ LDA \cite{blei2003latent} \\ NMF \cite{lee2000algorithms} \\ NeuralLDA \cite{Srivastava2017} \\ ProdLDA \cite{Srivastava2017} \\ ETM \cite{dieng2020topic} \\ DecTM \cite{wu2021discovering} \\ NSTM \cite{zhao2021nstm} \\ CombinedTM \cite{bianchi2021pre} \\ BERTopic \cite{grootendorst2022bertopic} \\ TSCTM \cite{wu2022mitigating} \\ ECRTM \cite{wu2023effective} \\ FASTopic \cite{wu2024fastopic} }}
    & \makecell[l]{TC \\ TD \\ Classification \\ Clustering}
    & \multirow{2}[3]{*}{\makecell[l]{20NG \\ IMDB \\ Wikitext-103 \\ NeurIPS \\ ACL \\ NYT}} \\
    \cmidrule{1-3}Hierarchical topic modeling
    & {\makecell[l]{HDP \cite{teh2006} \\ SawETM \cite{duan2021sawtooth} \\ HyperMiner \cite{xu2022hyperminer} \\ ProGBN \cite{duan2023bayesian} \\ TraCo \cite{wu2024traco}}}
    & \makecell[l]{TC over levels \\ TD over levels \\ Classification over levels \\ Clustering over levels}
    &  \\
    \midrule
    Dynamic topic modeling & {\makecell[l]{ DTM \cite{blei2006dynamic} \\ DETM \cite{dieng2019dynamic} \\ CFDTM \cite{wu2024dynamic} }} & \makecell[l]{TC over time slices \\TD over time slices \\Classification \\Clustering} & \makecell[l]{NeurIPS \\ ACL \\ NYT} \\
    \midrule
    Cross-lingual topic modeling & \makecell[l]{ NMTM \cite{Wu2020} \\ InfoCTM \cite{wu2023infoctm} } & \makecell[l]{TC (CNPMI) \\ TD over languages \\ Classification \\ Clustering} & \makecell[l]{ECNews \\ Amazon \\ Review Rakuten} \\
    \bottomrule
    \end{tabular}%
    \caption{
        Summary of topic modeling scenarios, topic models, evaluation metrics, and datasets covered by \topmost.
    }
    \label{tab_all_scenarios}
\end{table*}

\section{Overview of Toolkit Design and Architecture}
    In this section, we delineate the overview of our toolkit design and architecture.
    We build \topmost with Python and use PyTorch \cite{paszke2019pytorch} as the neural network framework for neural topic models.
    \Cref{fig_architecture} illustrates the overall architecture of \topmost.

    \subsection{Topic Modeling Scenarios and Topic Models}
        As summarized in \Cref{tab_all_scenarios},
        \topmost reaches a wider coverage by involving the 4 most popular topic modeling scenarios and their corresponding conventional or neural topic models.

        \paragraph{Basic Topic Modeling} discovers a number of latent topics from normal documents like news articles and web snippets,
        as the most common scenario \cite{blei2003latent}.
        For basic topic models,
        \topmost supports conventional LDA \cite{blei2003latent}, NMF \cite{lee2000algorithms},
        and most of the mainstream neural models such as ProdLDA \cite{Srivastava2017}, ETM \cite{dieng2020topic}, CombinedTM \cite{bianchi2021pre}, BERTopic \cite{grootendorst2022bertopic}, TSCTM \cite{wu2022mitigating}, ECRTM \cite{wu2023effective}, and FASTopic \cite{wu2024fastopic}.

        \paragraph{Hierarchical Topic Modeling} organizes topics into a tree structure instead of flat topics in the basic topic modeling \cite{griffiths2003hierarchical,isonuma2020tree}.
        Topics at each level of the structure involve different semantic granularity: child topics are more specific to their parent topics.
        This provides more desirable granularity for downstream applications.
        Hierarchical topic models in \topmost include conventional HDP \cite{teh2006} and recently popular neural hierarchical topic models, \eg HyperMiner \cite{xu2022hyperminer}, ProGBN \cite{duan2023bayesian}, and TraCo \cite{wu2024traco}.

        \paragraph{Dynamic Topic Modeling} discovers the evolution of topics in sequential documents,
        such as the conference papers published by year \cite{blei2006dynamic}.
        This discloses how topics emerge, grow, and decline over time due to real-world trends and events,
        which has derived applications like trend analysis and public opinion mining \cite{li2020global,churchill2022dynamic}.
        For dynamic topic models,
        we provide the conventional DTM~\cite{blei2006dynamic} and its neural variant, DETM~\cite{dieng2019dynamic}.
        We also cover recent CFDTM~\cite{wu2024dynamic}.

        \paragraph{Cross-lingual Topic Modeling} discovers aligned cross-lingual topics from bilingual corpora \cite{mimno2009polylingual}.
        These reveal the commonalities and differences across languages and cultures, enabling cross-lingual text analysis without supervision \cite{yuan2018multilingual,yang2019multilingual}.
        Cross-lingual topic models in \topmost include NMTM~\cite{Wu2020} and InfoCTM~\cite{wu2023infoctm}.

        We carefully adapt the original implementations of these topic models and unify their APIs of initialization, training, and testing,
        ensuring that our toolkit remains user-friendly, readable, and extendable.
        Note that we will constantly update \topmost to include more newly released models.

    \subsection{Datasets and Preprocessing}
        \topmost contains extensive benchmark datasets for the involved topic modeling scenarios, as reported in \Cref{tab_all_scenarios}.
        We summarize the statistics of these datasets in \Cref{tab_datasets_basic_hierarchical,tab_datasets_dynamic,tab_datasets_cross-lingual}.

        For basic and hierarchical topic modeling, we have the following datasets:
        \begin{inparaenum}[(i)]
            \item
                \textbf{20NG} \cite[20 News Groups, ][]{Lang95} is one of the most widely used datasets for evaluating topic models, including news articles with 20 labels.
            \item
                \textbf{IMDB}~\footnote{\url{http://ai.stanford.edu/~amaas/data/sentiment/aclImdb_v1.tar.gz}}
                \cite{maas2011learning} is the movie reviews from the IMDB website, containing two sentimental labels, positive and negative.
            \item
                \textbf{Wikitext-103}~\footnote{\url{https://www.salesforce.com/products/einstein/ai-research/the-wikitext-dependency-language-modeling-dataset/}} 
                \cite{merity2016pointer} includes Wikipedia articles \cite{nguyen2021contrastive}.
        \end{inparaenum}

        For dynamic topic modeling, \topmost provides the datasets as
        \begin{inparaenum}[(i)]
            \item
                \textbf{NeurIPS}~\footnote{\url{https://www.kaggle.com/datasets/benhamner/nips-papers}}
                includes the published papers at the NeurIPS conference from 1987 to 2017.
            \item
                \textbf{ACL}~\cite{bird2008acl}
                is an article collection between 1973 and 2006 from ACL Anthology
                ~\footnote{\url{https://aclanthology.org/}}.
            \item
                \textbf{NYT}~\footnote{\url{https://huggingface.co/datasets/Matthewww/nyt_news}}
                contains the news articles in the New York Times, from 2012 to 2022,
                with 12 categories, like ``Arts'', ``Business'', and ``Health''.
        \end{inparaenum}

        For cross-lingual topic modeling, we offer the following bilingual datasets:
        \begin{inparaenum}[(i)]
            \item
                \textbf{ECNews}~\footnote{\url{https://github.com/bobxwu/NMTM}}
                \cite{Wu2020} is a collection of English and Chinese news with 6 categories like business, education, and entertainment.
            \item
                \textbf{Amazon Review} \cite{Wu2020} includes English and Chinese reviews from the Amazon website, where each review has a rating from one to five.
                We simplify it as a binary classification task by labeling reviews with ratings of five as ``1'' and the rest as ``0'' following \citet{yuan2018multilingual}.
            \item
                \textbf{Rakuten Amazon} \cite{wu2023infoctm} contains Japanese reviews from Rakuten \cite[a Japanese online shopping website, ][]{zhang2017encoding}, and English reviews from Amazon \cite{yuan2018multilingual}. Similarly, it is also simplified as a binary classification task according to the ratings.
        \end{inparaenum}
        Note that basic topic models can employ the datasets for dynamic topic modeling as well.

        We preprocess these datasets with standard steps, such as removing stop words and punctuation, removing short tokens, and filtering low-frequency words \cite{Card2018a,Wu2020short}.
        Users can directly download these off-the-shelf datasets for experiments through \topmost from our GitHub repository.
        See \Cref{app_datasets} for more details of these datasets.
        We also provide configurable preprocessing implementations, allowing users to flexibly customize their datasets.

    \subsection{Evaluation Metrics}
        \topmost provides sufficient evaluation metrics to evaluate topic models.
        We first evaluate the quality of discovered topics in terms of \textbf{topic coherence} \cite[\textbf{TC}, ][]{Newman2010} and \textbf{topic diversity} \cite[\textbf{TD}, ][]{dieng2020topic}.
        TC refers to the coherence between the top words of discovered topics, and TD measures the differences between topics.
        We consider different implementations of TC and TD, for example, NPMI \cite{lau2014machine}, $C_V$ \cite{roder2015exploring}, and TU \cite{Nan2019}, for extensive comparisons.

        Then, we evaluate the quality of inferred doc-topic distributions via extrinsic tasks:
        \textbf{text classification} and \textbf{clustering} \cite{Wu2019,zhao2021nstm,nguyen2024topic}.
        For classification, we train an ordinary classifier (\eg SVM) with doc-topic distributions as document features and predict the labels of others.
        For clustering, we use the most significant topics in doc-topic distributions as clustering assignments.

        Apart from these fundamental ones, we additionally include metrics for special scenarios.
        For cross-lingual topic modeling,
        we measure the average TD over all languages and evaluate the alignment between cross-lingual topics with cross-lingual NPMI \cite[CNPMI, ][]{hao2018learning}.
        For hierarchical topic modeling, we evaluate the quality of discovered topic hierarchies,
        concerning the coherence and diversity between parent and child topics,
        the diversity between parent and non-child topics, and the diversity between sibling topics \cite{chen2021tree,chen2021hierarchical,wu2024traco}.

\section{Comparison to Existing Toolkit}
    To highlight our significant strengths, we compare our \topmost with the latest counterpart, OCTIS \cite{terragni2021octis},
    which integrates more features than earlier toolkits.
    Our \topmost outperforms OCTIS in three key aspects:
    \begin{enumerate}[wide, labelindent=0pt, itemsep=0pt, topsep=5pt, label=(\textbf{\roman*})]
        \item
            As detailed in \Cref{tab_comparison}, \topmost offers a broader coverage of topic modeling scenarios,
            accompanied by corresponding datasets, models, and evaluation metrics.
            This better fulfills the various requirements of researchers and developers.
        \item
            \topmost provides a more extensive array of topic models compared to OCTIS.
            As reported in \Cref{fig_venn_comparison} while OCTIS merely includes 4 neural topic models, \topmost incorporates 16 ones, including the latest NSTM \cite{zhao2021nstm}, HyperMiner \cite{xu2022hyperminer}, and ECRTM \cite{wu2023effective}.
            These advanced models empower users with cutting-edge topic modeling techniques and simplify their comparisons and applications.
        \item
            \topmost entirely decouples the implementations of data loading, model construction, model training, and evaluations, as illustrated in \Cref{fig_architecture}.
            This design streamlines the code structure for high reusability and facilitates fair comparisons among diverse topic models.
            It aligns with prominent libraries such as Huggingface Transformers and PyTorch Lightning.
            See the code examples in \Cref{sec_usage}.
    \end{enumerate}

\section{Toolkit Usage} \label{sec_usage}
    We showcase the simplicity and user-friendly design of our \topmost toolkit with code examples.
    Users can directly install our \topmost through pip~\footnote{\url{https://pypi.org/project/topmost}}: \texttt{pip install topmost}.

    \Cref{fig_code_quickstart} shows how to quickly utilize \topmost to discovers topics from documents
    with a few handy steps: dataset preprocessing, model construction (here ProdLDA \cite{Srivastava2017}), and training.
    We emphasize that our \topmost supports other languages besides English.
    We can simply employ different tokenizers in the preprocessing for other languages,
    for example, jieba~\footnote{\url{https://github.com/fxsjy/jieba}} for Chinese and nagisa~\footnote{\url{https://github.com/taishi-i/nagisa}} for Japanese.
    Other preprocessing settings are also configurable, including maximum vocabulary size, stop words, and maximum or minimum document frequency.
    This allows users to flexibly apply our toolkit.

    \begin{figure}[H]
        \vspace{-\intextsep}
        \begin{lstlisting}
from topmost.data import RawDatasetHandler
from topmost.models import ProdLDA
from topmost.trainers import BasicTrainer

docs = [ "A document about space, satellite, launch, orbit.", # more example documents...
]
# build a dataset
dataset = RawDatasetHandler(docs)
# create a topic model
model = ProdLDA(dataset.vocab_size)
# create a trainer
trainer = BasicTrainer(model)
topic_top_words, doc_topic_dist = trainer.fit_transform(dataset)
        \end{lstlisting}
        \caption{
            A code example for quick start.
        }
        \label{fig_code_quickstart}
        \vspace{-5pt}
\end{figure}

    \Cref{fig_code_training} exemplifies how to train a topic model with preprocessed datasets.
    The training of other topic models follows similar steps.

    \begin{figure}[H]
        \vspace{-\intextsep}
        \begin{lstlisting}
from topmost.data import download_dataset, BasicDatasetHandler
from topmost.models import ProdLDA
from topmost.trainers import BasicTrainer

#download a dataset
download_dataset('20NG', cache_path='./datasets')
# load a dataset
dataset = BasicDatasetHandler("./datasets/20NG")

# create a topic model
model = ProdLDA(dataset.vocab_size)
# create a trainer
trainer = BasicTrainer(model)
# train the topic model
trainer.train(dataset)
        \end{lstlisting}
        \caption{
            A code example for training a topic model
            (ProdLDA \cite{Srivastava2017}).
        }
        \label{fig_code_training}
\end{figure}

    \Cref{fig_code_evaluation} shows how to fully evaluate the trained topic model
    with diverse metrics including topic coherence, topic diversity, text classification, and text clustering.

    \begin{figure}[H]
        \vspace{-\intextsep}
        \begin{lstlisting}
from topmost.evaluations import compute_topic_diversity, compute_topic_coherence, evaluate_clustering, evaluate_classification

# doc-topic distributions
train_theta, test_theta = trainer.export_theta(dataset)
# top words of topics
topic_top_words = trainer.export_top_words(dataset.vocab)
# topic coherence
compute_topic_coherence(topic_top_words, dataset.train_texts)
# topic diversity
compute_topic_diversity(topic_top_words)
# text clustering
evaluate_clustering(test_theta, dataset.test_labels)
# text classification
evaluate_classification(train_theta, test_theta, dataset.train_labels, dataset.test_labels)
        \end{lstlisting}
        \caption{
            A code example for evaluating a topic model,
            including topic coherence, topic diversity, text classification, and clustering.
        }
        \label{fig_code_evaluation}
\end{figure}

\begin{figure*}[!ht]
    \centering
    \includegraphics[width=\linewidth]{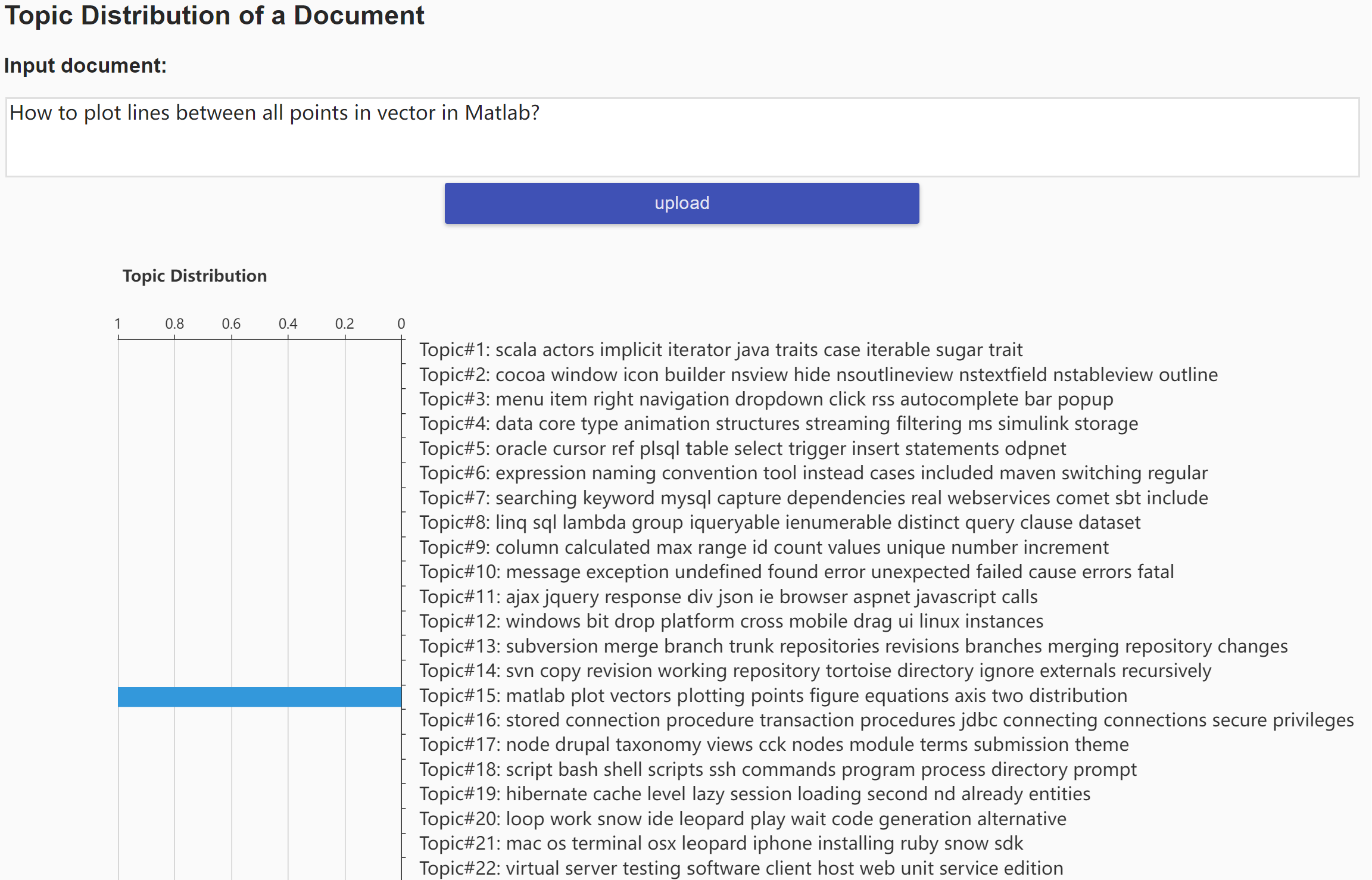}
    \caption{
        Demonstration of testing new documents.
        It plots the inferred topic distribution of an input document from a trained topic model.
    }
    \label{fig_demo_input}
\end{figure*}

\begin{figure}[!ht]
    \centering
    \includegraphics[width=\linewidth]{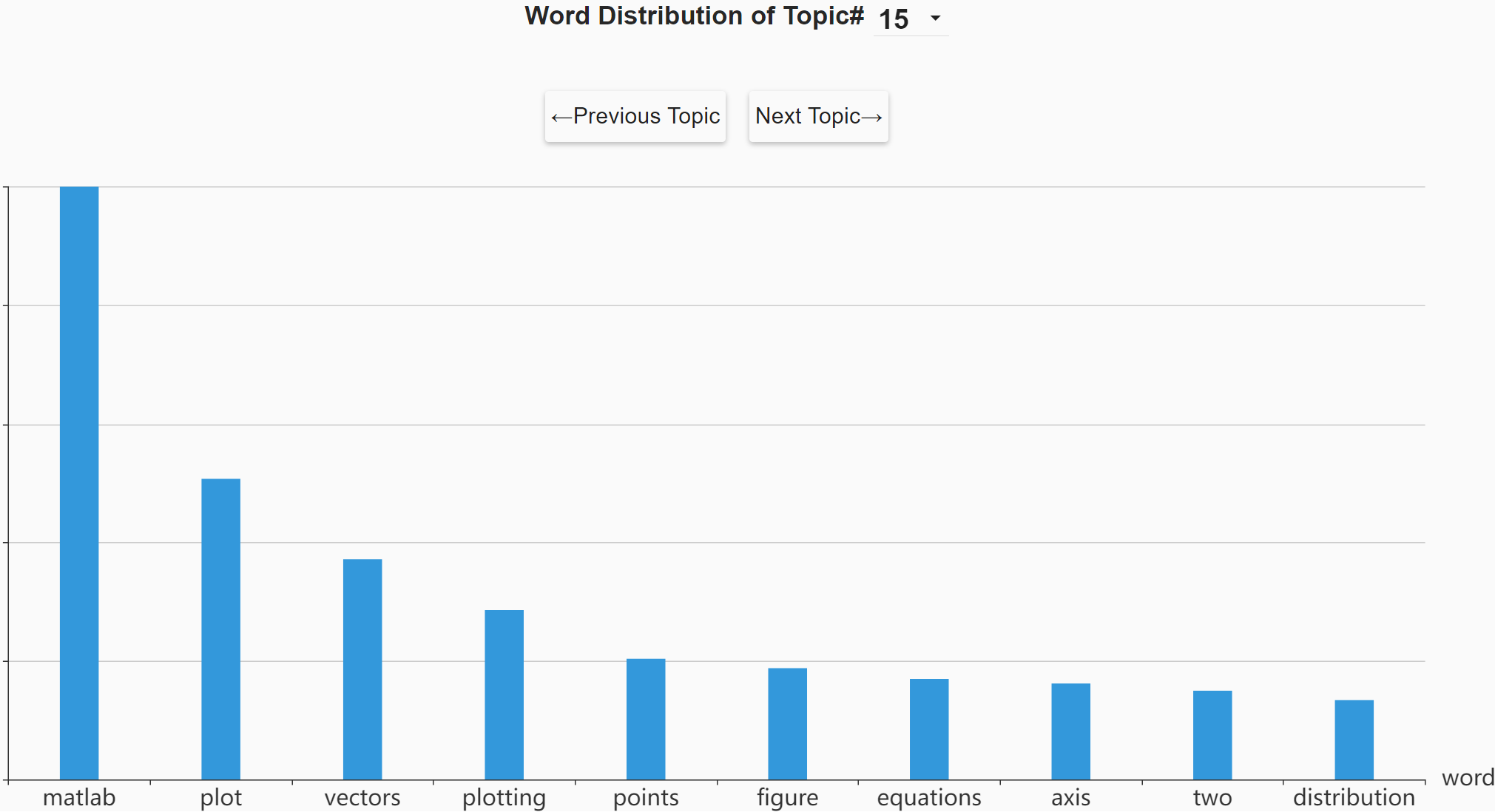}
    \caption{
        Visualization of discovered topics.
        It plots the top related words of each topic and the modeled word distributions.
    }
    \label{fig_demo_topics}
\end{figure}

    The above examples illustrate that \topmost clearly decouples the APIs of data, models, and training,
    so \topmost becomes more accessible, easy-to-use, and extendable to users.
    Due to limited page space, see more examples and tutorials on our GitHub project page,
    like data preprocessing and other topic modeling scenarios.

\section{Visualization Interfaces}
    \topmost furthermore provides visualization interfaces for topic models.
    We create a web demo system with Flask~\footnote{\url{https://flask.palletsprojects.com/}} as the server framework following Material design
    to visualize and test topic models.
    It is designed to be intuitive and user-friendly, enabling users to easily understand and leverage topic models.

    \Cref{fig_demo_topics} shows the visualization of topics.
    We plot the top related words of discovered topics and the modeled probability of each word.
    For example, Topic\#15 in \Cref{fig_demo_topics} mostly relates to words like ``matlab'', ``plot'', and ``vectors''.
    By selecting the index or clicking the \textit{Previous} and \textit{Next} buttons, we can view the details of any topic.

    \Cref{fig_demo_input} demonstrates the interactive utilization of a trained topic model.
    Upon inputting a document,
    we can click the \emph{upload} button to obtain the inferred topic distribution of the document.
    The horizontal bar chart in \Cref{fig_demo_input} plots the distribution over all topics of the input \textit{How to plot lines between all points in vector in Matlab?}.
    We see that the topic distribution mainly lies on Topic\#15, which refers to Matlab.

\section{Conclusion and Future Work}
    In this paper, we present \topmost, an open-source, comprehensive, and up-to-date topic modeling system toolkit.
    \topmost provides complete lifecycles of various topic modeling scenarios,
    including datasets, preprocessing, models, training, and evaluations,
    which outperforms existing counterparts.
    \topmost allows users to smoothly explore topic models, verify their new ideas, and develop novel topic modeling applications.
    This benefits both the communities in academia and industry.
    In the future, we plan to keep \topmost updated to incorporate more latest topic models
    and support more features to facilitate the research and application of topic modeling.

\section*{Limitations}
    We consider the following limitations of \topmost.
    First, \topmost only includes the mainstream evaluation metrics.
    Some less popular ones like perplexity are ignored.
    Second, \topmost does not cover the topic models based on prompting large language models \cite{pan2023fact,wu2024updating,pham2023topicgpt}.
    Different from LDA-like models, they define a topic is defined as a textual description, so we cannot assess them through existing evaluation metrics.

\bibliography{lib}

\clearpage
\appendix

\begin{table*}[!t]
    \centering
    \small
    \begin{tabular}{llrrrr}
        \toprule
        Dataset & Language & \#docs & \makecell[r]{Vocabulary \\ Size} & \makecell[r]{Average \\ length} & \#labels \\
        \midrule
        \multirow{2}[0]{*}{ECNews} & English &                  46,870  &                    5,000  & 12.0 & \multirow{2}[0]{*}{6} \\
              & Chinese &                  50,000  &                    5,000  & 10.6 &  \\
        \midrule
        \multirow{2}[0]{*}{Amazon Review} & English &                  25,000  &                    5,000  & 30.6  & \multirow{2}[0]{*}{2} \\
              & Chinese &                  25,000  &                    5,000  & 43.2  &  \\
        \midrule
        \multirow{2}[0]{*}{Rakuten Amazon} & English &                  25,000  &                    5,000  & 30.6  & \multirow{2}[0]{*}{2} \\
              & Japanese &                  25,000  &                    5,000  & 22.5  &  \\
        \bottomrule
    \end{tabular}%
    \caption{
        Statistics of pre-processed datasets for cross-lingual topic modeling.
    }
    \label{tab_datasets_cross-lingual}%
\end{table*}%

\begin{table}[!t]
    \centering
    \resizebox{\linewidth}{!}{
        \begin{tabular}{lrrrr}
        \toprule
        Dataset & \#docs & \makecell[r]{Vocabulary \\ Size} & \makecell[r]{Average \\ Length} & \#labels \\
        \midrule
        20NG  &        18,846  &                    5,000  &                    110.5  & 20 \\
        IMDB  &          50,000  &                    5,000  &                      95.0  & 2 \\
        Wikitext-103 &        28,532  &                  10,000  &                 1,355.4  & / \\
        \bottomrule
        \end{tabular}%
    }
    \caption{
        Statistics of pre-processed datasets for basic and hierarchical topic modeling.
    }
    \label{tab_datasets_basic_hierarchical}%
\end{table}%

\begin{table}[!t]
    \centering
    \resizebox{\linewidth}{!}{
        \begin{tabular}{lrrrrr}
        \toprule
        Dataset & \#docs & \makecell[r]{Vocabulary \\ Size} & \makecell[r]{Average \\ Length} & \#labels & \makecell[r]{\#time \\ slices} \\
        \midrule
        NeurIPS & 7,237  &                  10,000  &                    2,085.9  & /     & 31 \\
        ACL   & 10,560  &                  10,000  & 2,023.0 & /    & 31 \\
        NYT   & 9,172  &                  10,000  & 175.4 & 12    & 11 \\
        \bottomrule
        \end{tabular}%
    }
    \caption{
        Statistics of pre-processed datasets for dynamic topic modeling.
    }
    \label{tab_datasets_dynamic}%
\end{table}%

\section{Datasets} \label{app_datasets}
    \Cref{tab_datasets_basic_hierarchical,tab_datasets_dynamic,tab_datasets_cross-lingual} report the statistics of datasets for different topic modeling scenarios after preprocessing.
    Users can directly download all these datasets via \topmost from our GitHub repository.

\end{document}